%% file: acl_latex.tex
\pdfoutput=1

\documentclass[11pt]{article}

\usepackage{acl}

\usepackage{times}
\usepackage{latexsym}

\usepackage[T1]{fontenc}

\usepackage[utf8]{inputenc}

\usepackage{microtype}

\usepackage{inconsolata}
\usepackage[final]{pdfpages} 
\usepackage{booktabs}
\usepackage{multirow}
\usepackage{graphicx}
\usepackage{color}
\usepackage{makecell}
\usepackage{setspace}
\usepackage{enumitem}
\usepackage{tablefootnote}
\usepackage{hyperref}

\usepackage[most]{tcolorbox}

\usepackage{fontawesome}
\usepackage{tikz}
\usepackage{pgfplots}
\usepackage{subfigure}
\usepackage{graphicx}

\definecolor{block-gray}{gray}{0.85}
\definecolor{xlinkcolor}{cmyk}{1,0.6,0,0}

\makeatletter
\def\thanks#1{\protected@xdef\@thanks{\@thanks
        \protect\footnotetext{#1}}}
\makeatother
\title{

Clustering and Ranking: Diversity-preserved Instruction Selection \\through Expert-aligned Quality Estimation
}

\author{Yuan Ge$^{1*}$\thanks{$*$~\ Work done during an internship at Huawei.}, Yilun Liu$^2$\textsuperscript{\faEnvelope}\thanks{\textsuperscript{\scalebox{1.12}{\faEnvelope}}  Corresponding author (liuyilun3@huawei.com).}, Chi Hu$^1$, Weibin Meng$^2$, Shimin Tao$^2$, Xiaofeng Zhao$^2$,\\
\textbf{Hongxia Ma$^2$, Li Zhang$^2$, Boxing Chen$^3$, Hao Yang$^2$, Bei Li$^1$, Tong Xiao$^{1,4}$, Jingbo Zhu$^{1,4}$} \\
$^1$ Northeastern University, Shenyang, China \\ 
$^2$ Huawei, Beijing, China \\ 
$^3$ Huawei Canada, Toronto, Canada \\ 
$^4$ NiuTrans Research, Shenyang, China \\
}

\begin{document}
\maketitle
\begin{abstract}

With contributions from the open-source community, a vast amount of instruction tuning (IT) data has emerged. 
Given the significant resource allocation required for training and evaluating models, it is advantageous to have an efficient method for selecting high-quality IT data. 
However, existing methods for instruction data selection have limitations such as relying on fragile external APIs, being affected by biases in GPT models, or reducing the diversity of the selected instruction dataset.
In this paper, we propose an industrial-friendly, expert-aligned and diversity-preserved instruction data selection method: \textbf{C}lustering \textbf{a}nd \textbf{R}anking (\textbf{CaR}).
CaR employs a two-step process: first, it ranks instruction pairs using a high-accuracy (84.25\%) scoring model aligned with expert preferences; second, it preserves dataset diversity through clustering.
In our experiment, CaR efficiently selected a mere 1.96\% of Alpaca's IT data, yet the resulting AlpaCaR model surpassed Alpaca's performance by an average of 32.1\% in GPT-4 evaluations. Moreover, we find that data selecting is a consistent paradigm whether the pre-trained model is more capable or the model parameters scaling up. Our approach employs compact models with 550M parameters and incurs just 11.2\% of the financial outlay of current methods, enhancing its industrial deployability.

\end{abstract}

\section{Introduction}
Language Models (LMs) acquire the capability to follow instructions through Instruction Tuning (IT) \citep{radford2019language, brown2020language, zhang2023instruction}, which aligns Large Language Models (LLMs) with critical human standards such as security, privacy, and legal compliance. 
Self-instruct proposes a novel methodology that utilizes LMs to construct IT datasets \citep{wang2022self}, greatly improving the efficiency of instruction generation. Alpaca leveraged a similar strategy \citep{alpaca}, utilizing \texttt{text-davinci-003} to construct the Alpaca\_52k dataset, and subsequent IT on LLaMA-7B model \citep{touvron2023LLaMA} led to the creation of Alpaca.

\begin{figure}
	\centering
	\includegraphics[width=78mm]{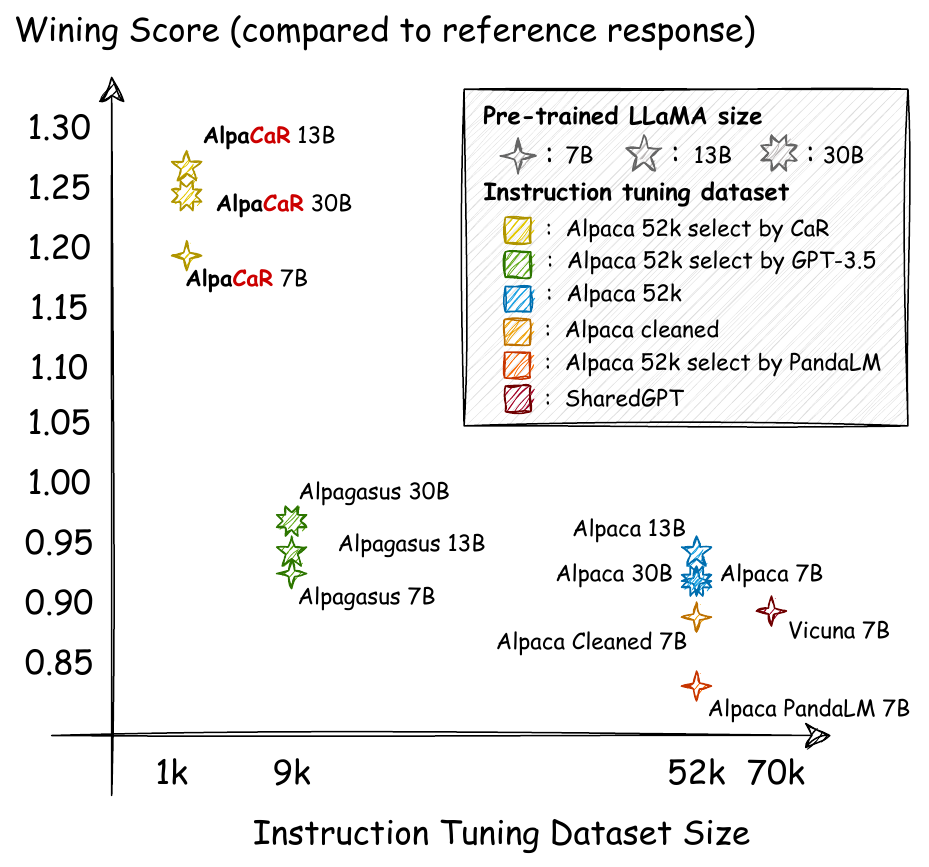}
	\caption{Compares the performance of the proposed AlpaCaR model to established baseline models over four test sets. Our AlpaCaR achieves the best model performance with the smallest amount of instruction tuning data.}
 \label{fig:main_result}
\end{figure}

Despite these advancements, the quality of instructions remains paramount over their quantity.
\citet{zhou2023lima} carefully curated 1,000 instructions, ensuring data quality and diversity by human being, resulting in LIMA model significantly outperforming the Alpaca. 
Nevertheless, creating high-quality instruction sets through manual annotation is both time-consuming and labor-intensive \citep{vicuna2023}. 
A promising approach to mitigate this challenge involves filtering a small subset of high-quality and diverse instructions from the vast amounts of existing instruction data.

Alpagasus \citep{chen2023alpagasus} introduced a straightforward yet effective method that utilizes \texttt{GPT-3.5-Turbo} to filter roughly 9k instructions, surpassing Alpaca's performance. However, this approach overlooks data diversity, and \texttt{GPT}'s evaluations rated 17.3\% instruction pairs generated by \texttt{text-davinci-003} above 4.5 and 74.9\% above 4.0, demonstrating GPT's self-enhancement bias \citet{zheng2023judging}, rendering it unsuitable for assessing instructions generated by models within the same series. 
Therefore, more authentic human preferences should be used to filter instruction sets. 
Moreover, relying on fragile and expensive external GPT APIs limits Alpagasus in industrial deployment, especially in low-computation resource scenarios.

\begin{table}
\centering
\resizebox{0.88\linewidth}{!} {%
\begin{tabular}{ccccc}
\toprule[0.7pt]
   \textbf{IQS} & Comet$_{Instruct}$  & GPT-4  & GPT-3.5 \\ \hline
   \textbf{84.25\%} & 72.44\% & 63.19\% & 57.48\% \\
   \textbf{78.12\%} & 45.00\% & 65.00\% & 56.25\% \\
\toprule[0.7pt]
\end{tabular}
}
\caption{\label{test set}
Accuracy of the IQS, Comet$_{Instruct}$ and GPT models on test sets. Reflecting the alignment of the model with human preferences in the task of Instruction Pairs Quality Estimation. The second row presents results for instruction pairs sourced from the \textit{IQE test set}, while the third row shows acc on instruction pairs from \textit{Vicuna\_80}, demonstrating the models' generalization to other distributions, see more details in Appendix \ref{sec:IQE}. The IQS and Comet$_{Instruct}$ model were fine-tuned as described in Appendix \ref{sec:IQS}, while the GPT model used prompts referenced in the Appendix \ref{sec:compare_prompt}.
}
\end{table}

In this work, we propose an effective and efficient method for selecting instruction pairs --- \textbf{C}lustering \textbf{a}nd \textbf{R}anking (\textbf{CaR}). CaR consists of two steps. The first is ranking through quality estimation on instruction pairs, where an expert-aligned scoring model (with 550M parameters only) achieves an accuracy of 84.25\% with expert preferences. 
Then, a clustering step ensures the overall diversity of the dataset, minimizing potential capability gaps. 
Our contributions are summarized as follows:
\begin{itemize}
    \item We introduce Instruction Pair Quality Estimation (IQE), a new stage before IT process which aims to use the assessment results of instruction datasets as an aid for the actual fine-tuning of language models and evaluation on benchmarks, reducing the time and computational expenses for model performance validation in IT process by over 90\%.
    \item We propose a novel quality evaluation paradigm for IT dataset that is independent of external APIs and aligns well with human experts' preferences.
    As shown in Table \ref{test set}, our small Instruction pair Quality Scoring (IQS) model, compared to GPT-4, achieves a 21.05\% improvement in aligning with human preferences for data quality.
    \item We propose CaR, an instruction selection method that aligns with expert insights and preserves diversity, showcasing significant enhancements in model performance and training efficiency.
    As shown in Fig. \ref{fig:main_result}, CaR uses a small model to filter high-quality instruction data, achieving an average performance exceeding Alpaca by about 13.3\% to 32.8\% on the Alpaca\_52k dataset using only a 1.96\% subset of instructions. This implies a reduction of 98\% in training time and resources. 
    \item In \hyperlink{sec:discussion}{section 5}, experiments found that the data selecting paradigm is effective even with \textit{more adequate pre-training} (LLaMA 1\textendash LLaMA 3) or \textit{model parameter scaling} (7B\textendash 30B). However, data selecting methods at \textit{higher data quality}, such as Alpaca-GPT4 \citep{peng2023instruction}, are still challenging.
\end{itemize}
In addition, we released our code and models to facilitate future research and industrial endeavors\footnote{\url{https://github.com/IronBeliever/CaR}}.
\section{Method}

\subsection{Motivation}

Our work is motivated by the challenges of data quality in
instruction tuning and the limitations of existing approaches.

\noindent
\paragraph{From Quality Estimation to Instruction Pair Quality Estimation.}
Quality estimation is a crucial task in machine translation (MT), enabling the assessment of MT models' effectiveness and the selection of high-quality translations for specific purposes, such as manual post-editing. Similarly, LLMs' IT process faces the challenge of rapidly shifting from rare to abundant instruction pairs with inconsistent quality. 
Ensuring the quality of IT datasets presents a significant challenge, necessitating adjustments to the pre-trained model, executing inference on test datasets, and undergoing evaluation by LLM or human annotators. These processes are not only time-intensive but also demand considerable computational resources. To address this, we propose a paradigm shift from evaluating model performance to assessing IT datasets via IQE. Our goal is to perform a coarse screening of a large number of instructions using IQE, followed by refining and selecting the optimal LLM with minimal datasets to reduce the overall computational cost associated with instruction filtering and verification.

\begin{figure*}[htbp]
    \centering
    \includegraphics[width=128mm]{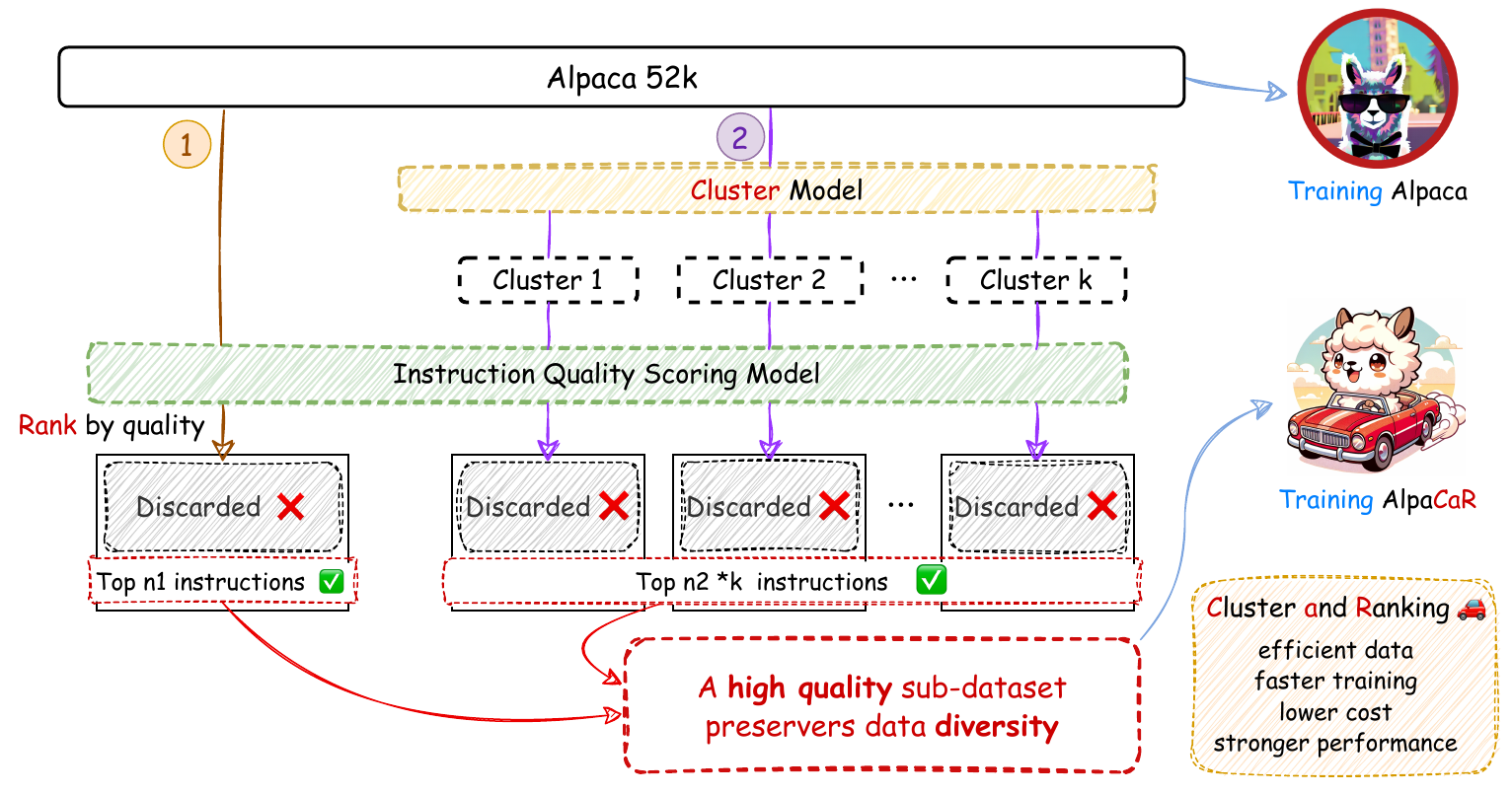}
    \caption{An overview of Cluster and Ranking (CaR) method. Unlike directly training Alpaca with the entire Alpaca\_52k dataset, CaR first uses the IQS model to score all instructions (\textcolor{brown}{brown arrow}). Then it selects the top $n_1$ instructions ranked by quality. Next, a clustering model (\textcolor{violet}{violet arrow}) groups all instructions into k clusters, selecting $n_2$ from each. These are concatenated and deduplicated to form a diverse, high-quality sub-dataset for training AlpaCaR.}
    \label{fig:main_method}
\end{figure*}

\noindent
\paragraph{GPT as a Judge Exhibits Systematic Bias.}
Researchers often use GPT preferences as a proxy for human preferences in scenarios requiring human feedback, due to time and cost considerations \citep{zhou2023lima, rafailov2023direct, dubois2023alpacafarm, lee2023rlaif}. However, \texttt{GPT-4} has been shown to exhibit systemic biases in its evaluations, including positional bias, verbosity bias, and self-enhancement bias \citep{zheng2024judging, wang2023large}.
While researchers generally view Alpaca 52k as needing improvement (AlpacaDataCleaned \footnote{\url{https://github.com/gururise/AlpacaDataCleaned}} ; \citealp{liu2023automatic}),  GPT's evaluations rated 9k instruction pairs above 4.5 and 39k above 4.0. 
Introducing more realistic human preferences for instruction filtering could further enhance model performance.

\noindent
\paragraph{Instruction Diversity Inspires LLMs' Multi-tasks Capability.}
Recent studies have highlighted the importance of data diversity in improving the performance of LLMs \citep{zhou2023lima, chen2023alpagasus}.  
\citet{dong2023abilities} found that combining training data from various tasks boosts LLMs' performance in low-resource scenarios.
Inspired by these findings, we posit that integrating instructions from different tasks enhances LLMs' capabilities in low-resource settings. Consequently, ensuring the diversity of the IT dataset is paramount, particularly when dealing with large-scale models and limited high-quality data for each task.

\subsection{Clustering and Ranking Method}\label{sec:CAR_method}
Considering the aforementioned motivations, we propose a straightforward yet effective data selection framework, Cluster and Ranking, which integrates the dimensions of quality and diversity.
Inspired by \citet{zhou2023lima}'s work, we first select a subset that ensures the retention of a large number of high-quality instructions, then supplement a small number of high-quality instructions from each cluster to enhance data diversity while preserving instruction quality.
As illustrated in Fig. \ref{fig:main_method}, the framework begins by evaluating the entire dataset using the IQS model, assigning a $score_i$ to each instruction $pair_i$. 
Subsequently, the cluster model is employed to partition all candidate instruction pairs into $k$ clusters. 
Finally, all instruction pairs are sorted based on their scores, and the top $n_1$ pairs are selected; Within each cluster, the top $n_2$ pairs are chosen based on their scores.
The resulting high-quality sub-dataset with preserved diversity is curated by deduplicating $n_1+k*n_2$ pairs of instructions and is intended for the training of AlpaCaR.

Sections \ref{sec:quality} and \ref{sec:diversity} provide a comprehensive discussion of the ranking and clustering methodologies implemented in CaR.

\subsection{Single Instruction Pair Quality Estimation}\label{sec:quality}

To explore the IQE task, we adapt the Comet framework \citep{rei-etal-2020-comet} and develop a suitable framework for leveraging expert preference.
Our training data is derived from expert-revised dataset \citep{liu2023automatic}, consisting of 3,751 instruction pairs from Alpaca\_52k that were refined by linguistic experts to enhance fluency, accuracy, and semantic coherence between questions and responses. 
We categorize unedited instructions and responses from \texttt{text-davinci-003} as \textit{GPT Preference}, and expert-revised instructions as \textit{Expert Preference}. To enable the model to discern features across these categories, we curated 2,541 markedly distinct instructions from the expert-revised dataset, ensuring an edit distance above a small threshold. These instruction pairs 
are then randomly allocated them into training, validation, and test sets following an 8:1:1 distribution.

Initially, we experimented with the translation ranking model architecture from the Comet framework to leverage the paired annotations in expert-revised better. In Fig. \ref{fig:IQS&Comst} (left), Comet$_{instruct}$ optimizes the model using instruction and input as anchors, minimizing semantic distance to human-preferred responses while maximizing distance to GPT-generated outputs.
This approach achieves 72.44\% accuracy on the test set but fails to fully leverage the improvements about \textit{Input} made by experts.
To address this, as illustrated in Fig. \ref{fig:IQS&Comst} (right), we retained the pre-trained XLM-RoBERTa large in Comet$_{instruct}$ and directly concatenated the instruction pair components to train the IQS model.
As shown in Table \ref{test set}, our IQS model outperforms \texttt{GPT-3.5} (version: \texttt{GPT-3.5-Turbo}) and \texttt{GPT-4} (version: \texttt{GPT-4-1106-preview}). Further analysis reveals that \texttt{GPT-4} favors original instructions in 62.2\% of incorrect cases, showing that even advanced GPT models often prefer GPT-aligned instructions. Additionally, \texttt{GPT-4} struggles to recognize nuanced semantic changes made by experts in 37.8\% of incorrect cases, revealing its difficulty in recognizing expert and nuanced semantic changes with minimal adjustments.
Despite \texttt{GPT-4}'s strong alignment with human preferences in most general tasks, its subpar performance on the expert-revised dataset highlights a subtle gap between expert preferences and GPT preferences.

\subsection{Diversity}\label{sec:diversity}
Within the instruction filtering framework, it is imperative to filter out a minimal subset of data from a vast array of instructions, resulting in a limited number of instructions per task. In such low-resource scenarios, 
\citet{dong2023abilities} has demonstrated that blending training data from various tasks enhances the LLMs' proficiency across different abilities.
Intuitively, by assigning a task label to each instruction pair, we can preserve instruction pairs associated with a broader range of tasks, thereby facilitating cross-task instruction synergy and enhancing model performance. To determine task labels for instruction pairs, we evaluated manual labeling, classification models, and clustering models, selecting clustering for our study. Manual labeling, though more accurate, is labor-intensive and less adaptable to various datasets. We hypothesize that instruction pairs within the same task are semantically close, allowing their distribution to be learned via classification models. Nonetheless, such models may struggle with flexibility when faced with out-of-domain data.

To enhance the method's versatility, we opted for an unsupervised clustering-based approach to preserve data diversity. A clustering algorithm can identify semantically close instruction pairs and form clusters for different tasks. Moreover, this choice allows for efficient adaptation to different datasets without retraining from scratch by forming new clusters when encountering out-of-domain instruction pairs. 

Regarding the clustering methodology, we employ the $k$-Means algorithm.
Initially, a sentence-transformers model is used to map sentences to a 384-dimensional dense vector space. Subsequently, semantic features are PCA-reduced to retain 95\% of dimensions. Finally, by setting the number of clusters as $k=\sqrt{n/2}$, all 52k instruction pairs are clustered into 161 clusters. The diversity of the instruction sub-dataset is maintained by adjusting the quantity of instruction pairs within each cluster.

\begin{table*}
\centering
\resizebox{\linewidth}{!}{ 
\begin{tabular}{lccccccccccccccc}\toprule
\multicolumn{1}{l}{\multirow{2}*{{\textbf{Method}}}} & \multicolumn{1}{l}{\multirow{2}*{{\textbf{Num}}}} & \multicolumn{1}{l}{\multirow{2}*{{\textbf{Size}}}}  & \multicolumn{3}{c}{{\textbf{PandaLM}}}  &\multicolumn{3}{c}{{\textbf{Vicuna}}} &\multicolumn{3}{c}{{\textbf{CoachLM}}} &\multicolumn{3}{c}{{\textbf{Self-instruct}}}\\
\cmidrule(r){4-6}
\cmidrule(r){7-9}%
\cmidrule(r){10-12}%
\cmidrule(r){13-15}%
\multicolumn{3}{c}{~} & WS$^\uparrow$ & WR$^\uparrow$ & QS$^\uparrow$ & WS$^\uparrow$ & WR$^\uparrow$ & QS$^\uparrow$ & WS$^\uparrow$ & WR$^\uparrow$ & QS$^\uparrow$ & WS$^\uparrow$ & WR$^\uparrow$ & QS$^\uparrow$ \\ \midrule

Alpaca-PandaLM & 52k & 7B & 1.224  & 49.4\% & 72.9\% & 0.288  & 8.8\% & 20.0\% & 0.867  & 28.7\% & 58.0\% & 1.075  & 42.9\% & 64.7\% \\ 
Alpaca-cleaned & 52k & 7B & 1.276  & 53.5\% & 74.1\% & 0.300  & 8.8\% & 21.3\% & 0.953  & 35.3\% & 60.0\% & 1.083  & 42.5\% & 65.9\% \\ 
Vicuna & 70k & 7B & 1.276  & 53.5\% & 74.1\% & 0.688  & 17.5\% & 51.3\% & 0.787  & 23.3\% & 55.3\% & 0.877  & 25.8\% & 61.9\% \\ 
Alpaca & 52k & 7B & 1.341  & 54.1\% & 80.0\% & 0.363  & 11.3\% & 25.0\% & 0.913  & 32.7\% & 58.7\% & 1.139  & 42.9\% & 71.0\%  \\
Alpagasus & 9k & 7B & 1.324  & 54.1\% & 78.2\% & 0.463  & 13.8\% & 32.5\% & 0.807  & 25.3\% & 55.3\% & 1.123  & 44.4\% & 67.9\% \\ 
\textbf{AlpaCaR} & 1k & 7B & \textbf{1.594}  & \textbf{70.6\%} & \textbf{88.8\%} & \textbf{0.813}  & \textbf{27.5\%} & \textbf{53.8\%} & \textbf{1.020}  & \textbf{37.3\%} & \textbf{64.7\%} & \textbf{1.448}  & \textbf{61.9\%} & \textbf{82.9\%} \\ \midrule
Alpaca & 52k & 13B & 1.365  & 56.5\% & 80.0\% & 0.363  & 8.8\% & 27.5\% & 0.940  & 30.7\% & 63.3\% & 1.155  & 45.2\% & 70.2\% \\ 
Alpagasus & 9k & 13B & 1.347  & 54.7\% & 80.0\% & 0.338  & 6.3\% & 27.5\% & 0.880  & 28.0\% & 60.0\% & 1.230  & 48.4\% & 74.6\% \\ 
\textbf{AlpaCaR} & 1k & 13B & \textbf{1.535}  & \textbf{65.9\%} & \textbf{87.6\%} & \textbf{1.025}  & \textbf{37.5\%} & \textbf{65.0\%} & \textbf{1.153}  & \textbf{44.0\%} & \textbf{71.3\%} & \textbf{1.357}  & \textbf{56.3\%} & \textbf{79.4\%} \\ \midrule
Alpaca & 52k & 30B & 1.276  & 50.0\% & 77.6\% & 0.425  & 11.3\% & 31.3\% & 0.900  & 28.0\% & 62.0\% & 1.155  & 43.7\% & 71.8\% \\ 
Alpagasus & 9k & 30B & 1.382  & 57.1\% & 81.2\% & 0.438  & 8.8\% & 35.0\% & 0.920  & 30.0\% & 62.0\% & 1.214  & 46.8\% & 74.6\% \\ 
\textbf{AlpaCaR} & 1k & 30B & \textbf{1.553}  & \textbf{67.1\%} & \textbf{88.2\%} & \textbf{0.950}  & \textbf{28.8\%} & \textbf{66.3\%} &\textbf{ 1.120}  & \textbf{43.3}\% & \textbf{68.7}\% & \textbf{1.377}  & \textbf{57.1\%} & \textbf{80.6\%} \\ 

\bottomrule
\end{tabular}
}
\caption{\label{main experiment} 
Comparative analysis of AlpaCaR and existing methods in the primary experiment. Winning rates are determined relative to the reference responses of the test sets, providing a quantitative measure of performance.
} 
\end{table*}

\section{Experimental Setup}
To compare AlpaCaR with other models, we obtain a single response for each test set sample using a fixed prompt \citep{alpaca}. Judge LLMs are then compare responses generated by LLMs against each other or human reference responses, identifying their preferred responses. PandaLM, GPT-4 and human are used as judge, yielding consistent evaluation conclusions.

\subsection{Test Datasets}
To avoid confusion arising from the similarity in naming between models and datasets, we use the format ``ModelName\_DatasetSize'' to represent datasets. 
Following previous methodologies, we assess four datasets: Self-instruct\_252 \citep{li2023self}, Vicuna\_80 \citep{vicuna2023}, PandaLM\_170 \citep{wang2023pandalm}, and CoachLM\_150 \citep{liu2023automatic}. This approach covers a broader range of instructions, minimizing evaluation bias.

\subsection{Generations}
For each test instruction, a single response is generated from each baseline model using LLaMA-Factory's default settings~\cite{zheng-etal-2024-llamafactory}: temperature=0.95, top\_p=0.7, top\_k=50, no beam search, and a maximum token length to 512.

\subsection{Evaluate Metrics}
For each sample, the judge model receives a single instruction and two candidate responses. It labels the winning response or a tie if both stand out significantly. To address potential bias of LLM judges preferring specific positions, we tested the results twice by swapping the response order and define the final judgment based on:
\begin{itemize}[itemsep=2pt,topsep=0pt,parsep=0pt]
\item \textit{win} : win twice, or win once and tie once 
\item \textit{lose} : lose twice, or lose once and tie once
\item \textit{tie} : tie twice, or win once and lose once
\end{itemize}

We compute three types of winning rates: (1) WS, a winning score formulated as WS=$1+\frac{\#win-\#lose}{\#all}$.
(2) WR, which considers wins cases and is given by WR= $\frac{\#win}{\#all}$, where $\#all$ is the number of test set samples; 
(3) QS, a quality score that measures the ratio of responses reaching the reference level, formulated as QS= $\frac{\#win+\#tie}{\#all}$.

Evaluation Approach:
(1) GPT-4 Turbo, currently the most powerful LLM widely used to replace manual responses quality assessments, with prompts designed by \citet{vicuna2023}. However, this method faces limitations due to API dependency and inherent biases.
(2) PandaLM, an open-source evaluation model that can be deployed locally, providing efficient LLM assessments \citep{wang2023pandalm}. Trained on 300k samples using GPT-3.5, it effectively mitigates biases and achieves 88.3\% of GPT-4's evaluation capability.
(3) Human, three experts with an average of 12.57 years of experience independently conducted comparisons based on the criteria in Appendix \ref{sec:evaluation criteria}
After comprehensive consideration, we use the evaluation results of PandaLM to measure the model's instruction-following ability in most experiments, while some key principal experiments utilize GPT-4 and human for assessment.
The prompt for GPT-4's evaluation is designed by \citet{vicuna2023}, as detailed in the Appendix \ref{sec:eval_prompt}.

\section{Results and Analysis}
In this section, we compared AlpaCaR with baseline models, including Alpaca, Alpaca-PandaLM, Alpaca-cleaned, Alpagasus, and Vicuna. We replicated all baseline models at a 7B scale and demonstrated the superiority of AlpaCaR at 13B and 30B scales.

\subsection{Comparison with Baselines}
We conduct a comparative analysis of two established baseline LLMs, Alpaca and Vicuna, which were fine-tuned using 52,000 text instructions through text-davinci-003 and 70,000 ChatGPT dialogues, respectively. Furthermore, we explore three models that advance upon Alpaca: Alpaca-PandaLM and Alpaca-cleaned, which employ instructional enhancement methods, and Alpagasus, which incorporates an instruction filtering method. All models were trained with identical hyperparameter settings. As delineated in Table \ref{main experiment}, AlpaCaR, at the 7B scale, outperforms not only the foundational models of Alpaca and Vicuna but also Alpaca-PandaLM, Alpaca-cleaned, and Alpagasus. Overall, AlpaCaR achieves significant performance improvements over Alpaca across the 7B, 13B, and 30B scales, validating the efficacy of the CaR method. The notable performance gains of AlpaCaR, accomplished with reduced data usage compared to Alpagasus, underscore the importance of leveraging high-quality human preferences and data diversity in enhancing model performance.

\input{latex/figures/fig3}

\subsection{Reliability of IQE Results}
To verify whether the IQE results genuinely reflect the performance of LLMs after IT, we examined the correlation between scores given by the IQS model and the performance of fine-tuned LLMs on test sets. Given that Alpagasus obtained 9k instructions rated above 4.5 using \texttt{GPT-3.5-Turbo}, we similarly selected the top 9k instructions ranked by IQS model and Comet model. We then calculated the average score for the three IT sub-datasets using the IQS model, fine-tuned LLaMA-7B, and tested its performance by averaging models' winning scores on four datasets against reference. As illustrated in Fig. \ref{fig:IQE-LLM}, the average IQS score and the fine-tuned model's performance are generally consistent, indicating that IQE results can approximately reflect the performance of LLMs after fine-tuning.

\input{latex/figures/fig4/fig4}

\subsection{Ablation Study}
\paragraph{Quality Dimension.}
To illustrate the significance of data quality, we employed the IQS model's score to rank 52,000 instructions. Subsequently, we extracted subsets of the top 1,000, 2,000 and up to 42,000 instructions to train LLaMA-7B. In Fig. \ref{fig:quality}, 
the horizontal axis represents the size of instruction dataset, where a higher count signifies more instructions of relatively lower quality, while the vertical axis shows the winning score relative to Alpaca.
The results indicate that models trained with selected data generally surpass the one trained with the entire dataset. As more instructions of relatively lower quality are included, the performance of the LLM generally declines. 
Remarkably, the model approaches its optimal performance with a mere 1,000 high-quality IT data. Therefore, in the CaR method, we select $n_1=1000$ instructions to ensure the chosen IT sub-dataset is of high quality.

\input{latex/figures/fig5/fig5}

\paragraph{Selection of $n_2$: Trade-off between Quantity and Quality.}
We compared the number of samples selected from each cluster after $k$-means clustering. Fig. \ref{fig:cluster_result} demonstrates that, compared to using only 1k high-quality data selected by IQS model, the CaR method enhances performance when a small number of samples (up to 5) are selected from each cluster. Selecting too many samples can negatively impact the overall quality of the IT sub-dataset and the performance of the LLMs. Moreover, the CaR method achieves nearly optimal performance by selecting $n_2=1$ sample from each cluster, thus enhancing the diversity of the IT sub-dataset.

\paragraph{Importance of Diversity.}
An ideal IT dataset should encompass a rich variety of data, but determining the optimal number of instructions per cluster required for the model to effectively correspond to the task remains a challenge. 
We designed experiments to demonstrate the importance of diversity and explore values of $n_2$, the trade-off between the number and quality of samples per cluster.

Designing strict ablation experiments in this context is challenging due to the difficulty in ensuring consistent instruction set quality while maintaining the same number of instructions. To explore this, we established three experimental groups with increasing diversity (baseline: reference response). In Table \ref{Diversity ablation}, the winning rates on the Self-Instruct and Vicuna test sets show that models with more diverse instruction sets perform better.

\begin{table}[tp]
\centering
\resizebox{\linewidth}{!}{ 
\begin{tabular}{lcccccc}\toprule
\multicolumn{1}{l}{\multirow{2}*{{\textbf{Method}}}}  & \multicolumn{3}{c}{{\textbf{Vicuna}}} &\multicolumn{3}{c}{{\textbf{Self-instruct}}}\\
\cmidrule(r){2-4}
\cmidrule(r){5-7}%
    & WS$^\uparrow$ & WR$^\uparrow$ & QS$^\uparrow$ & WS$^\uparrow$ & WR$^\uparrow$ & QS$^\uparrow$ \\ \hline
    $40\times4$  & 0.625  & 20.0\% & 31.3\%  & 1.226  & 48.4\% & 61.3\%  \\ 
    $80\times2$  & 0.600  & 18.8\% & 30.0\%  & 1.290  & 52.4\% & 64.5\%  \\ 
    $160\times1$  & 0.688  & 23.8\% & 34.4\%  & 1.365  & 59.5\% & 68.3\%  \\ 
\bottomrule
\end{tabular}
}
\caption{\label{Diversity ablation} Ablation on Diversity: Models with more diverse instruction sets perform better. ($160\times1$ means 1 highest IQS-scored sample per 160 clusters)} 
\end{table}

\input{latex/figures/fig6}

\input{latex/figures/fig7}

\subsection{Compare with Random \& GPT-4 Result}
Fig. \ref{fig:method_ablation} presents the results of ablation experiments, revealing that randomly selecting 1,017 instruction pairs from 52k dataset leads to a decrease in model performance compared to Alpaca. In contrast, the instruction pairs selected by the CaR method show significant improvements at 7B (29.8\%), 13B (32.7\%), and 30B (33.1\%) scales.

Furthermore, to address cost considerations, we employed GPT-4's evaluation framework exclusively on four datasets to compare AlpaCaR against Alpaca. As depicted in Fig. \ref{fig:gpt-4_result_vicuna} and elaborated upon in Appendix \ref{sec:gpt-4_result}, GPT-4 exhibited similar evaluative outcomes: AlpaCaR outperformed baseline in the majority of instances, thereby substantiating the efficacy of the CaR method. Employing CaR, which involves selecting 1.96\% of the dataset, has proven to yield superior preferences across a variety of parameter scales.

\subsection{Human Evaluation}
We have formulated detailed evaluation criteria, covering seven aspects: fluency, relevance, correctness, consistency, satisfaction, informativeness and security, which are further categorized into 27 primary and 58 secondary classifications. Additional details are provided in Appendix \ref{sec:evaluation criteria}.

We compared AlpaCaR 30B vs. Alpaca 30B on Vicuna\_80 test set. The human evaluation results demonstrated that AlpaCaR performed at least as well as Alpaca across all categories and was preferred by language experts in the vast majority of cases. The specific results are shown in Table \ref{human eval}.

\begin{table}[t]
    \centering
    \small
    \begin{tabular}{lcccc}
    \toprule[0.7pt]
        \textbf{Category} & \textbf{win} & \textbf{lose} & \textbf{tie} & \textbf{WS$^\uparrow$}  \\ \hline
        Writing & 8 & 1 & 1 & 1.700  \\ 
        Roleplay & 5 & 0 & 5 & 1.500  \\ 
        Common-sense & 9 & 0 & 1 & 1.900  \\ 
        Fermi & 7 & 2 & 1 & 1.500  \\ 
        Counterfactual & 7 & 0 & 3 & 1.700  \\ 
        Coding & 3 & 3 & 1 & 1.000  \\ 
        Math & 0 & 0 & 3 & 1.000  \\ 
        Generic & 6 & 0 & 4 & 1.600  \\ 
        Knowledge & 7 & 2 & 1 & 1.500  \\ \hline
        \textbf{Total} & 52 & 8 & 20 & \textbf{1.550}  \\ 
        \toprule[0.7pt]
    \end{tabular}
    \caption{\label{human eval}Human evaluation results on Vicuna\_80 dataset: AlpaCaR\_30B vs. Alpaca\_30B.}
\end{table}

Table \ref{case study} in Appendix \ref{sec: case study} displays \textit{case study} from the math category. We found that under strict evaluation criteria, experts believed that neither model provided the correct final answer, resulting in a tie. However, a more detailed analysis reveals that \textit{AlpaCaR utilized CoT to explore the correct reasoning steps}, although errors occurred after certain steps. In contrast, Alpaca simply provided a confusingly incorrect answer.
We hypothesize that the IQS model has learned experts' preferences for detailed reasoning processes presented in the training data. Consequently, during subset selection, the IQS model favors instruction pairs that showcase meticulous reasoning, resulting in the fine-tuned AlpaCaR exhibiting more comprehensive thought processes in the form of CoT reasoning.

\subsection{Larger Instruction Tuning Datasets}
To further explore the performance of CaR in more massive and complex datasets, we conducted additional experiments on even larger instruction datasets. Following recent work \citep{du2023mods, liu2023makes}, we combined five instruction tuning datasets, including Alpaca, Dolly\_v2 \citep{conover2023free}, Alpaca-evol-instruct \citep{xu2023wizardlm}, HC3 \citep{guo2023close}, and LIMA \citep{zhou2023lima}, to obtain a large-mixed-dataset containing 181,253 instructions. Then we used CaR to filter the large-mixed dataset and obtained CaR\_50k containing 50k instructions.

\begin{table}[tp]
\centering
\resizebox{\linewidth}{!}{ 
\begin{tabular}{lcccccc}\toprule
\multicolumn{1}{l}{\multirow{2}*{{\textbf{Method}}}}  & \multicolumn{3}{c}{{\textbf{Vicuna}}} &\multicolumn{3}{c}{{\textbf{Self-instruct}}}\\
\cmidrule(r){2-4}
\cmidrule(r){5-7}%
    & WS$^\uparrow$ & WR$^\uparrow$ & QS$^\uparrow$ & WS$^\uparrow$ & WR$^\uparrow$ & QS$^\uparrow$ \\ \hline
    Alpaca & 0.338 & 10.00\% & 16.88\% & 1.206 & 45.63\% & 60.32\% \\ 
    mixed-181k & 0.875 & 28.80\% & 43.75\% & 1.349 & 52.38\% & 67.46\% \\ 
    CaR\_50k & 1.113 & 33.75\% & 55.62\% & 1.500 & 63.89\% & 75.00\% \\ 
\bottomrule
\end{tabular}
}
\caption{\label{large mixed} CaR is a stable and effective framework even on larger datasets} 
\end{table}

Table \ref{large mixed} shows that the model fine-tuned on 50k instructions selected by CaR outperforms Alpaca at the same number of instructions using LLaMA 2 7B as the base pre-trained model. In addition, the model fine-tuned using CaR\_50k outperforms the one using mixed-181k instruction tuning dataset.

This illustrates that the bottleneck of Alpaca is not that pre-trained LLaMA cannot learn more knowledge from more instructions, but rather that the limited quality of instruction dataset restricts the model's performance. It also demonstrates that CaR is a stable and effective framework even on larger datasets. CaR framework can filter 50k high-quality instructions from 181k instruction pairs to get stronger model performances with less training overheads.

\begin{table}[t]
    \centering
    \small
    \begin{tabular}{lccc}
    \toprule[0.7pt]
        Method & Selection & Training & Total \\ \hline
        Alpaca & $0\$$ & $733.35\$$ & $733.35\$$ \\ 
        Alpagasus & $12.66\$$ & $104.18\$$ & $116.84\$$ \\ 
        AlpaCaR & $0.02\$$ & $13.07\$$ & $ 13.09\$$ \\ 
    \toprule[0.7pt]
    \end{tabular}
    \caption{\label{cost comparison}
    Cost comparison of 30B scale.}
\end{table}

\subsection{Cost Comparison}
Here, we compare the computational costs of AlpaCaR, Alpaca, and Alpagasus, focusing on instruction evaluation and full parameter fine-tuning at the 30B scale, as detailed in Table \ref{cost comparison}.
For instruction evaluation using an API-based method, we refer to the official pricing \footnote{\url{https://openai.com/pricing}}, while for model training or inference, we consider the rental costs of GPUs
\footnote{\url{https://www.leadergpu.com/}}.
In summary, training AlpaCaR significantly saves both time and costs, compared to Alpaca or Alpagasus.

\section{Is the Benefit Derived from Data Selecting Universally Applicable?}
\label{sec:discussion}
Filtering a high quality instruction sub-dataset to supervised fine-tuning LLaMA 1 significantly reduces computational cost and effectively improves LLM performances. 
More crucially, it is essential to ascertain whether data screening constitutes a consistent paradigm for performance enhancement, particularly as pre-trained model become increasingly powerful and model parameters scaling up. In this section, we used the average WS on Vicuna\_80 and Self-instruct\_252 test set to explore the generalization of data selection.

\paragraph{A consistent paradigm when \textit{pre-training is more adequate}?} 
Base pre-trained LLMs acquire knowledge through pre-training. LLaMA 1, LLaMA 2, and LLaMA 3 were pre-trained using 1T, 2.4T, and 15T tokens, respectively. When pre-trained models exhibit strong capabilities, can they discern the quality of fine-tuning instructions, rendering instruction selecting redundant?
To investigate this, we employed LLaMA 1 7B, LLaMA 2 7B, and LLaMA 3 8B pre-trained models, comparing fine-tuning using the full dataset or subsets filtered by GPT-3.5 Turbo or CaR. Fig. \ref{fig:pre-trained} shows the results on Alpaca\_52k and Dolly\_15k IT datasets.
The findings suggest that even as base pre-trained LLMs become more powerful, models fine-tuned on filtered data surpass those trained on full instructions. LLaMA 3 8B is more susceptible to low-quality instructions, impeding its ability to follow instructions in downstream tasks.

\input{latex/figures/fig8/fig8}

\input{latex/figures/fig9/fig9}

\paragraph{A consistent paradigm when \textit{model size scaling up}?}
Many new capabilities and phenomena emerge as the model parameters scaling up.
Thus another question is whether instruction tuning data selection is still important as the parameters increase. 
We experimented the performance of the model fine-tuned by full versus selected instructions at the 7B-30B scale, due to limited computational conditions.
As shown on the left side of Fig. \ref{fig:param&data} (left), The horizontal direction showed no significant improvement in model performance even as the model size increased. 
However, the vertical direction showed that the model performs better using instructions selected by GPT-3.5 or CaR at all scales.

\paragraph{A consistent paradigm when \textit{instructions quality improves}?}
Alpaca-GPT4 \citep{peng2023instruction} contains instruction generated by GPT-4 using Alpaca prompts, which quality significantly improved compared to Alpaca.
Distinguishing high-quality instructions remains a challenge when instruction quality generally improves. As depicted in Fig. \ref{fig:param&data} (right), models trained by CaR-selected instructions are inferior to full instructions. We argue that the IQS model cannot significantly discriminate instruction quality in such a high-quality data distribution, so randomly filtering instructions caused performance degradation similar to Fig. \ref{fig:method_ablation}.
A similar phenomenon occurs when using LLMs to select instructions. Qwen1.5-110B-chat and Qwen-max scored more than 1,800 of the 2,000 instructions in the Alpaca-GPT4 dataset as perfect score, indicating that the quality of the evaluated instructions in this situation approaching the boundaries of the LLMs' capabilities.
So data selecting methods at \textit{higher data quality} are still challenging, and maybe gradient-based \citep{xia2024less} or in-context learning-based \citep{li2023one} methods demonstrate greater potential.

\section{Conclusion}
In this paper, we focus on exploring and resolving the issue of instruction selection during supervised fine-tuning stage. We introduce the CaR method and examine two perspectives that are warrant considered:
(1) Evaluating instruction quality using more authentic human preferences: models trained with data annotated by linguistic experts show higher agreement rates and the selected instructions lead to better-performing models.
(2) Instruction diversity inspires LLMs' stronger capability: Under our selection framework, preserving a small number of instructions for different tasks through cluster improves model performance. 
Experimental results show that fine-tuning LLaMA (ranging from 7B to 30B parameters) with a 1.96\% subset of instructions selected by CaR outperforms models trained on full datasets or data selected by GPT. Moreover, data selecting methods using GPT-family or CaR is a consistent paradigm whether the pre-trained model is more capable or the model parameters scaling up, while those at higher data quality are still challenging. Additionally, our approach can be deployed locally without relying on APIs, thereby enabling a more efficient instruction selection approach in low-computation resource environments.

\newpage

\section{Limitation}
Despite the outstanding performance of CaR across multiple test sets, its experiments were confined to filtering on only several datasets. The diverse formats of different open-source instruction sets pose challenges for the academic community interested in instruction filtering tasks. In the future, we plan to validate the effectiveness of CaR on more datasets such as WizardLM\_evol\_instruct\_70k \citep{xu2023wizardlm}. Moreover, while CaR is primarily used for single-turn dialogue instruction filtering, exploring its application in multi-turn dialogue instruction filtering presents an attractive direction for future research.

\section{Potential Risk \& Ethical Consideration}
We reveal the following potential risks of our research based on ethical considerations:

\begin{enumerate}
\item Quality of instruction data: While the proposed method aims to select high-quality instruction data, there is still a risk that the selected subset may not fully represent the diversity and complexity of the entire dataset. This could potentially lead to biased or incomplete training of models and cause adverse social impact.
\item Bias and fairness: As with any AI research, there is a need to ensure fairness and mitigate biases. The selection process and scoring model used in CaR should be carefully monitored to prevent any unintentional biases, such as favoring certain types of instructions or excluding underrepresented groups.
\item Industrial deployment and responsible use: As the method is designed for industrial scenarios, it is important to consider the responsible use of the developed models. Ensuring that the models are not used for unethical purposes or harmful applications is crucial. Additionally, monitoring and addressing any unintended consequences or biases that may emerge during deployment should be a priority.
\end{enumerate}

\section{Acknoledgement}
This work was supported in part by the National Science Foundation of China (No.62276056), the Natural Science Foundation of Liaoning Province of China (2022-KF-16-01), the Fundamental Research Funds for the Central Universities (Nos. N2216016 and N2316002), the Yunnan Fundamental Research Projects (No. 202401BC070021), and the Program of Introducing Talents of Discipline to Universities, Plan 111 (No.B16009).

\bibliography{cite}

\appendix

\section{Related work}
\paragraph{Quality Estimation and Comet framework.}
Quality estimation is a pivotal task in machine translation, involving scoring or ranking translation results to select higher-quality data. Comet \citep{rei-etal-2020-comet} leverages input and reference translations to accurately assess translation quality, employing two architectures: the Estimator model and the Translation Ranking model. The Estimator model directly predicts quality scores for each evaluation instance, while the Translation Ranking model learns parameters from paired evaluation data to predict reasonable quality scores.

\paragraph{Algorithm - Data Lifecycle.}
In the modern era of deep learning, high-quality data has become the cornerstone for training robust and effective models. Over the past decade, there has been a growing emphasis on the collection and curation of superior data \citep{chu2016data, motamedi2021data}. The emergence of data-centric AI has underscored the belief that data quality is as crucial as algorithmic advancements within the AI/ML lifecycle \citep{hajij2021data, zha2023data}. This paradigm shift has been particularly evident since the introduction of the Transformer architecture \citep{vaswani2017attention}, which has revolutionized the field of language modeling. Rather than focusing on disruptive innovations in model structure, researchers have concentrated on leveraging the effectiveness of the Transformer architecture by stacking transformer blocks to create more potent models. Additionally, significant improvements in model performance have been achieved through the construction of task-specific datasets and the enhancement of data quality \citep{zhou2023lima, chen2023alpagasus, li2023one}.

\paragraph{Futher perspective of clustering and ranking.}
Many domains have employed methods similar to clustering and ranking. In information retrieval, Google extensively utilizes the PageRank algorithm \citep{page1999pagerank} to calculate the importance of hyperlinks between webpages. Liu et al. developed a cluster-based retrieval model by constructing language models for clusters \citep{liu2004cluster}, combining documents within the same cluster and searching/ranking clusters based on query generation likelihood. Tang et al. enhanced the Bi-encoder's performance in dense information retrieval tasks by using clustering algorithms to generate "pseudo-query embeddings" \citep{tang2021improving}. Selecting suitable data for LLM inference is crucial in the RAG field, as discussed by \citet{yuan2023evaluating} and \citet{mu-etal-2023-augmenting}, who explore methods for finding appropriate demonstrations to improve LLM performance. In the network domain, Sun et al. introduced the RankClus framework \citep{sun2009rankclus}, which integrates clustering and ranking methods to strengthen heterogeneous information network analysis.

\paragraph{Evaluation of LLMs.}
Evaluating the open-domain instruction-following capabilities of LLMs presents a significant challenge. Currently, the prevailing approach involves employing human evaluators or GPT-4 to compare the inference response of different models. Consequently, recent studies, including PandaLM \citep{wang2023pandalm}, Vicuna \citep{vicuna2023}, CoachLM \citep{liu2023automatic}, and Self-Instruct \citep{wang2022self}, have curated and provided their own instruction sets to evaluate instruction-finetuned LLMs. Additionally, leaderboards such as MT-Bench \citep{zheng2024judging}, Alpaca-Eval \citep{dubois2023alpacafarm}, and Chatbot Arena \citep{chiang2024chatbot} have been established to measure the instruction-following abilities of these models.
PandaLM \citep{wang2023pandalm} and Auto-J \citep{li2023generative} efforts focus on training LLMs to provide more impartial and accurate evaluations. By leveraging these latest advancements, we aim to evaluate our model's performance using human-generated instruction sets, ensuring a comprehensive and rigorous assessment of its capabilities in following open-ended instructions.

\section{Evaluate Prompts}

\subsection{IQE Prompt}
\label{sec:eval_prompt}
\small
\newtcolorbox{mybackground}{colback=block-gray,grow to right by=0mm,grow to left by=0mm,boxrule=0pt,boxsep=0pt,breakable}
\begin{mybackground}
\noindent [The Start of Assistant A’s Instruction and Answer]

\noindent \{Instruction pair 1\}

\noindent [The End of Assistant A’s Instruction and Answer]

\noindent [The Start of Assistant B’s Instruction and Answer]

\noindent \{Instruction pair 2\}

\noindent [The End of Assistant B’s Instruction and Answer]

\noindent [System]

\noindent We would like to request your feedback on the performance of two AI assistants in response to the user question displayed above. Please rate the helpfulness, relevance, accuracy, level of details of their responses. Each assistant receives an overall score on a scale of 1 to 10, where a higher score indicates better overall performance. Please first output a single line containing only two values indicating the scores for Assistant 1 and 2, respectively. The two scores are separated by a space. In the subsequent line, please provide a comprehensive explanation of your evaluation, avoiding any potential bias and ensuring that the order in which the responses were presented does not affect your judgment.
\end{mybackground}

\subsection{Response Comparison Prompt}
\label{sec:compare_prompt}

\small
\begin{mybackground}
\noindent[Question]

\noindent \{Instruction\}

\noindent [The Start of Assistant 1's Answer]

\noindent \{Response 1\}

\noindent [The End of Assistant 1's Answer]

\noindent [The Start of Assistant 2's Answer]

\noindent \{Response 2\}

\noindent [The End of Assistant 2's Answer]

\noindent [System]

\noindent Please act as an impartial judge and evaluate the quality of the responses provided by two AI assistants to the user question displayed below. You should choose the assistant that follows the user’s instructions and answers the user’s question better. Your evaluation should consider factors such as the helpfulness, relevance, accuracy, depth, creativity, and level of detail of their responses. Begin your evaluation by comparing the two responses and provide a short explanation. Avoid any positional biases and ensure that the order in which the responses were presented does not influence your decision. Do not allow the length of the responses to influence your evaluation. Do not favor certain names of the assistants. Be as objective as possible. After providing your explanation, output your final verdict by strictly following this format: “[[A]]” if assistant A is better, “[[B]]” if assistant B is better, and “[[C]]” for a tie.
\end{mybackground}

\section{Specifics about Instruction Quality Estimation}

\subsection{Evaluation Metric of IQE}
\normalsize
\label{sec:IQE}

The second row of Table \ref{test set} presents results for instruction pairs sourced from the IQE test set, which are instructions revised  by language expert.
The third row shows accuracy on instruction pairs from Vicuna\_80, demonstrating the models' generalization to other distributions. The instructions are provided by the dataset, while language experts evaluates the quality of two responses generated by different models, establishing the ground truth labels.
In the calculation of accuracy, if the absolute difference between the scores of two responses is less than 0.01 assigned by IQS or Comet$_{Instruct}$, the outcome is considered a ``Tie''.

\subsection{Model Architecture of IQS and Comet$_{instruct}$}
\label{sec:IQS}

\begin{figure*}[tp] 
	\centering
	\includegraphics[width=120mm]{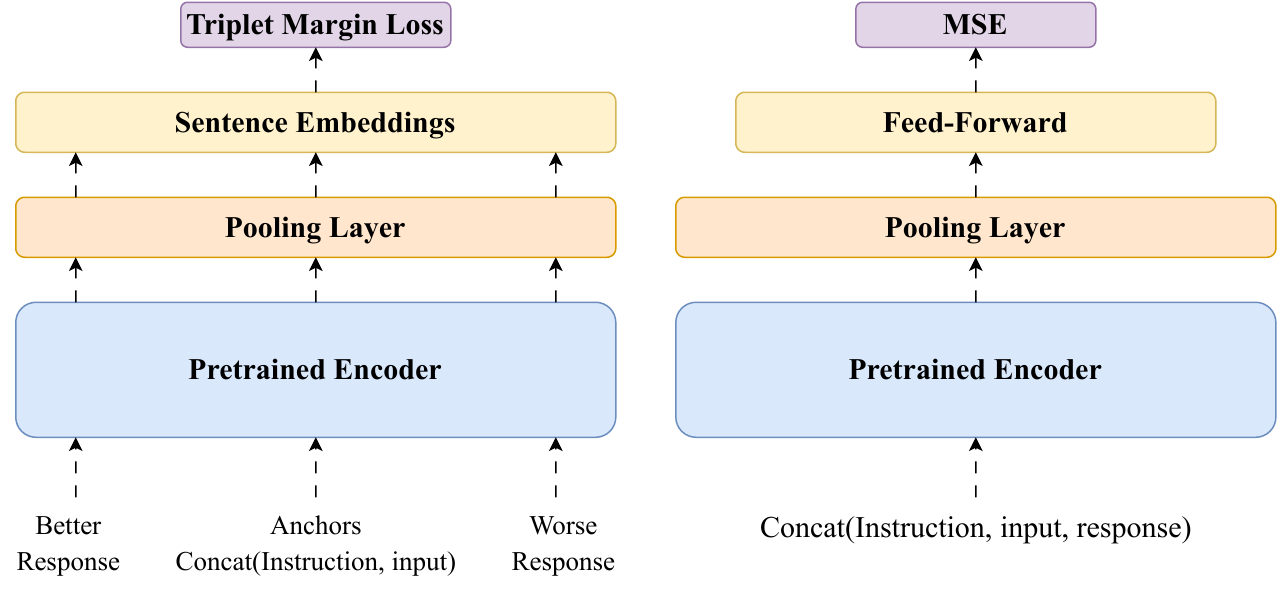}
	\caption{Detailed architecture of Comet$_{instruct}$ model(left) and Instruction pair quality scoring model(right).}
 \label{fig:IQS&Comst}
\end{figure*}

In the IQE task, the IQS model and Comet model correspond to the Estimator model architecture and Translation Ranking model architecture in the Comet framework, respectively. As shown in Fig. \ref{fig:IQS&Comst}, The Comet$_{instruction}$ model concatenates instructions with input to form anchors. It then feeds pairs of better and worse responses into the model. Finally, the model is trained using a triplet margin loss function to distinguish between the superior and inferior responses. The IQS model concatenates instruction pairs and then trains the model using Mean Squared Error as the loss function.

\section{More Results about GPT-4 Evaluations}
\label{sec:gpt-4_result}

As illustrated in Fig. \ref{fig:gpt4_CoachLM}, \ref{fig:gpt4_Sinstruct}, \ref{fig:gpt4_Pandalm}. Baseline: reference responses.

\input{latex/figures/fig11}

\input{latex/figures/fig12}

\input{latex/figures/fig13}

\section{Specifics about Human Evaluation Criteria}
\label{sec:evaluation criteria}

\small

\begin{itemize}
    \setlength{\parskip}{-0.1em}
    \item \textbf{Fluency}
    \begin{itemize}
        \item Redundancy: verbose repetition (>=2)
        \item Redundancy: extraneous content clutter
        \item Missing: incomplete response
        \item Error: syntax \& semantics issues
        \item Errors: word \& symbol errors
    \end{itemize}
    \item \textbf{Relevance}
    \begin{itemize}
        \item Irrelevant 
        \item Incorrect understanding of object / personality / relationship
        \item Poor compliance
    \end{itemize}
    \item \textbf{Correctness}
    \begin{itemize}
        \item Incorrect answer to the question
        \item Incorrect derivation process
    \end{itemize}
    \item \textbf{Consistency}
    \begin{itemize}
        \item Intrasentence consistency
        \item Contextual consistency
        \item Content Logical Consistency
        \item Query response consistency
    \end{itemize}
    \item \textbf{Satisfaction}
    \begin{itemize}
        \item Unclear views, no precise conclusions given (advice / recommendations)
        \item Insufficiently organized and comprehensive content
        \item Poor extensibility, almost a carbon copy of the query information, lack of embellishments
        \item Lack of emotional feedback
        \item Not careful enough to reply
        \item Biases in characterization and speaker tone
        \item Overly subjective and unsubstantiated
        \item Some of the recommendations are ineffective, irrelevant to the issue or lack specific advice and assistance
        \item Indeed a problem, but a subjective experience problem that is hard to describe
        \end{itemize}
        \item \textbf{Informativeness}
    \begin{itemize}
        \item Missing interpretive information
        \item Lack of richness
    \end{itemize}
    \item \textbf{Security}
    \begin{itemize}
        \item Minor or moderate security violations
        \item Serious security violations
    \end{itemize}
\end{itemize}

\newcommand{\tabincell}[2]{\begin{tabular}{@{}#1@{}}#2\end{tabular}}  
\begin{table*}[tp]
    \centering
    \small
    \begin{tabular}{ccccc}
    \hline
        \textbf{Instruction} & \textbf{Input} & \textbf{AlpaCaR\_30B} & \textbf{Alpaca\_30B} &  \makecell[l]{\textbf{Human} \\\textbf{evaluation}}  \\ \hline
        \makecell[l]{Given that f(x) = 5x\^{}3 - \\2x + 3, find the value \\of f(2).} & $\backslash$ & \makecell[l]{The value of f(2) = 5(2)\^{}3 - 2(2) + 3 \\= 5(8) - 4 + 3 = 44.} & The value of f(2) is 3. & Tie  \\ \hline
        \makecell[l]{Solve for x in the \\equation 3x + 10 = 5(x \\- 2).} & $\backslash$ & \makecell[l]{The equation 3x + 10 = \\5(x - 2) can be rearranged to solve for x. \\First, subtract 5(x - 2) from both sides \\to get 3x - 5x + 10 = 5(x - 2) - 5x + 10. \\Then, x = 5.} & x = 7 & Tie  \\ \hline
        \makecell[l]{If the endpoints of a \\line segment are (2, \\-2) and (10, 4), what is \\the length of the \\segment?} & $\backslash$ & \makecell[l]{The length of the line segment can be found \\using the formula for the distance between \\two points: \\d = $\surd$(x2 - x1)2 + (y2 - y1)2\\In this case, x2 = 10, x1 = 2, y2 = 4, and y1 = -2. \\Plugging these numbers into the formula, we get \\d = $\surd$(10 - 2)2 + (4 - (-2))2\\Similarly, we can calculate the length of the line \\segment using the Pythagorean-\\Pythagorean-Circles Theorem: \\d = $\surd$(x2 + y2)\\In this case, x2 = 10, y2 = 4, so \\d = $\surd$(10 + 4) = 5.\\Therefore, the length of the line segment is 5.} & \makecell[c]{The length of the \\line segment is 8.} & Tie  \\ \hline
    \end{tabular}
    \caption{\label{case study}Case study of LLMs responses in vicuna\_80 math category.}
\end{table*}

\normalsize
\section{Case study}
\label{sec: case study}
As illustrated in Table \ref{case study}.

\section{Profile of Involved Language Experts}

To ensure a comprehensive and rigorous human evaluation of LLM abilities, we established a collaboration with the language service center of a prominent international corporation. We recruited a team of highly educated, multilingual language experts with diverse skills in translation, localization, writing, and testing, who dedicated their full-time efforts to this task. Specifically, three experts possessing an average experience of 12.57 years, are responsible for conducting a human evaluation of AlpaCaR and other LLMs.

\section{Discussion of CaR framework}

Selecting top-n ranked samples for each cluster is indeed an intuitive and interesting idea that integrates the two steps of clustering and ranking. We have also experimented with this setting in our early research. However, a challenge arises when the predefined number of clusters
$ k = \sqrt{Number_{instructions/2}}=161$ is used. When top-n is small, the resulting dataset size is insufficient for the model to achieve good instruction-following capacity. Conversely, when top-n is large, it introduces more low-quality instruction pairs, which negatively impacts the performance of LLMs. An early version of our experimental results (baseline: Alpaca 52k) is shown in Table \ref{framework}.

\begin{table}[tp]
\centering
\resizebox{\linewidth}{!}{ 
\begin{tabular}{lcccccc}\toprule
\multicolumn{1}{l}{\multirow{2}*{{\textbf{Top-n}}}}  & \multicolumn{3}{c}{{\textbf{Vicuna}}} &\multicolumn{3}{c}{{\textbf{Self-instruct}}}\\
\cmidrule(r){2-4}
\cmidrule(r){5-7}%
    & WS$^\uparrow$ & WR$^\uparrow$ & QS$^\uparrow$ & WS$^\uparrow$ & WR$^\uparrow$ & QS$^\uparrow$ \\ \hline
    10 & 1.188   & 55.00\% & 90.00\%  & 1.230   & 45.63\% & 77.38\%  \\ 
    20 & 1.375   & 51.25\% & 83.75\%  & 1.167   & 42.86\% & 73.81\%  \\ 
    30 & 1.300   & 57.50\% & 85.00\%  & 1.111   & 38.49\% & 72.62\%  \\ 
    \textbf{CaR(ours)} & \textbf{1.475} & \textbf{58.75\%} & \textbf{88.75\%}  &  \textbf{1.310}  & \textbf{51.98\%} & \textbf{78.97\%}  \\ 
\bottomrule
\end{tabular}
}
\caption{\label{framework} Discussion of CaR framework: k $\times$ top-n v.s. $n_1$ + k $\times n_2$} 
\end{table}

The experimental results indicate that this combinatorial approach performs less effectively than treating the two components separately. Our idea is to additionally and separately extract the top $n_1$ instructions using only the ranking step to ensure that most high-quality instructions are included (as indicated in section \ref{sec:CAR_method}) while using a smaller top $n_2$ to prevent the inclusion of a large number of low-quality instruction pairs. Experimenting with different values of k might alleviate this problem, but we aim to propose a more automated process and avoid involving additional hyperparameter tuning.

\end{document}

%% file: latex/figures/fig3.tex
\definecolor{color2}{RGB}{098,190,166}
\definecolor{color3}{RGB}{253,186,107}
\definecolor{color4}{RGB}{235,096,070}
\begin{figure}[t]
\begin{tikzpicture}
\footnotesize{
\begin{axis}[
at={(0,0)},
    ymajorgrids,
    grid style=dashed,
    width=0.28\textwidth, height=0.25\textwidth,
    ybar=10pt,
    enlarge x limits=0.5,
    xtick align=inside,
    bar width=1.8em,
    symbolic x coords={{1}},
    nodes near coords align={vertical},
    ymin=0.7, ymax=1.3,
    xticklabels = {\scriptsize{Alpaca-Comet}, \scriptsize{Alpagasus}, \scriptsize{Alpaca-IQS} },
    xticklabel style={font=\scriptsize, rotate=15, anchor=east,xshift=1.5em,yshift=-0.7em},
    xlabel=\footnotesize{LLMs Performence},
    xlabel style={yshift=0.1em,align=center},
    ytick={0.8, 1.0, 1.2},
    yticklabels={0.8, 1.0, 1.2},
    scaled ticks=false,
    ]
    \addplot[fill=color2!50, draw=color2!80, area legend] coordinates {({1},0.879)};
    \addplot[fill=color3!50, draw=color3!80, area legend] coordinates {({1},0.929)};
    \addplot[fill=color4!50, draw=color4!80, area legend] coordinates {({1},1.183)};
\end{axis}
}
\hspace{3.8cm}
\footnotesize{
\begin{axis}[        
at={(3.8,0)},
    ymajorgrids,
    grid style=dashed,
    width=0.28\textwidth, height=0.25\textwidth,
    ybar=10pt,
    enlarge x limits=0.5,
    xtick align=inside,
    bar width=1.8em,
    symbolic x coords={{2}},
    nodes near coords align={vertical},
    ymin=0.2, ymax=0.8,
    xticklabels = {\scriptsize{Comet}, \scriptsize{GPT}, \scriptsize{IQS} },
    xlabel=\footnotesize{Average IQS Score},
    xlabel style={yshift=0.1em,align=center},
    ytick={0.3, 0.5, 0.7},
    yticklabels={0.3, 0.5, 0.7},
    scaled ticks=false,
    ]
    \addplot[fill=color2!50, draw=color2!80, area legend] coordinates {({2},0.378 )};
    \addplot[fill=color3!50, draw=color3!80, area legend] coordinates {({2},0.425)};
    \addplot[fill=color4!50, draw=color4!80, area legend] coordinates {({2},0.735)};

\end{axis}
}
\end{tikzpicture}

    \caption{Consistency between IQS scores and the performance of LLMs.}
    \label{fig:IQE-LLM}
\end{figure}

%% file: latex/figures/fig4/fig4.tex
\definecolor{color1}{RGB}{075,102,173}
\definecolor{color2}{RGB}{098,190,166}
\definecolor{color3}{RGB}{253,186,107}
\definecolor{color4}{RGB}{235,096,070}
\begin{figure}[t]
\centering
\begin{tikzpicture}
\centering
\footnotesize{
\begin{axis}[
at={(0,0)},
    ymajorgrids,
    grid style=dashed,
    width=0.45\textwidth, height=0.35\textwidth,
    legend columns=-1,
    legend style={fill opacity=0.5,text opacity =1, draw=none,line width=1pt, at={(0,1.25)}, anchor=north west, nodes={scale=0.8, transform shape} },
    xmin=0, xmax=50,
    ymin=0.85, ymax=1.25,
    xtick={0,10,20,...,50},
    ytick={0.9, 1.0, 1.1, 1.2},
    yticklabels={0.9, 1.0, 1.1, 1.2},
    ylabel=\footnotesize{Wining score relative to Alpaca},
    xlabel=\footnotesize{Size of IT dataset /k},
    xlabel style={yshift=0.1em,align=center},
    scaled ticks=false,
    ]
    \addplot[color1, mark=*, line width=1pt] file {latex/figures/fig4/data.txt}; 
    \addplot[red, dashed, line width=1pt] coordinates {(0,1)(50,1)}; 
    \node at (110, 100) [anchor=south] {baseline: Alpaca}; 
\end{axis}
}
\end{tikzpicture}

    \caption{Model performances with varying $n_1$.}
    \label{fig:quality}
\end{figure}

%% file: latex/figures/fig5/fig5.tex
\definecolor{color1}{RGB}{075,102,173}
\definecolor{color2}{RGB}{098,190,166}
\definecolor{color3}{RGB}{253,186,107}
\definecolor{color4}{RGB}{235,096,070}
\begin{figure}[t]
\centering
\begin{tikzpicture}
\centering
\footnotesize{
\begin{axis}[
at={(0,0)},
    ymajorgrids,
    grid style=dashed,
    width=0.45\textwidth, height=0.35\textwidth,
    legend columns=-1,
    legend style={fill opacity=0.5,text opacity =1, draw=none,line width=1pt, at={(0,1.25)}, anchor=north west, nodes={scale=0.8, transform shape} },
    xmin=0, xmax=20,
    ymin=1.16, ymax=1.36,
    xtick={0,5,10,...,20},
    ylabel=\footnotesize{Wining score relative to Alpaca},
    xlabel=\footnotesize{Number of samples selected from each cluster},
    xlabel style={yshift=0.1em,align=center},
    ytick={1.20, 1.25, 1.30, 1.35},
    yticklabels={1.20, 1.25, 1.30, 1.35},
    scaled ticks=false,
    ]
    \addplot[color1, mark=*, line width=1pt] file {latex/figures/fig5/data.txt}; 
    \addplot[red, dashed, line width=1pt] coordinates {(0,1.229431606)(20,1.229431606)};
    \node at (170, 70) [anchor=south] {baseline: 1k}; 
\end{axis}
}
\end{tikzpicture}
    \caption{Performances with varying $n_2$.}
    \label{fig:cluster_result}

\end{figure}

%% file: latex/figures/fig6.tex
\definecolor{ugreen}{RGB}{098,190,166}
\definecolor{uyellow}{RGB}{253,186,107}
\definecolor{ured}{RGB}{235,096,070}

\begin{figure}[t]
\centering
\pgfplotsset{
    width=0.5\textwidth,
   height=0.25\textheight,
   symbolic x coords={1,2,3},
   enlarge y limits={upper,value=0.05},
   legend style={
      fill,
      at={(0,16.5em)},
      legend columns=3,
      legend cell align=left,
      anchor=south
      },
   }
\footnotesize{
\begin{tikzpicture}

\begin{axis}[
    at={(0.42em,1em)},
    ymajorgrids,
    grid style=dashed,
    legend entries={Alpaca, Random, AlpacaCaR},
    legend cell align={left},
    ybar,
    enlarge x limits=0.28,
    xtick align=inside,
    bar width=1.4em,
    xmax=3,
    xmin=1,
    xtick=data,
    nodes near coords align={vertical},
    ymin=0.60,
    ymax=1.35,
    ytick={0.6,0.8,1.0,1.2},
    yticklabels={0.6,0.8,1.0,1.2},
    xticklabels={7B,13B,30B},
    xtick style={draw=none},
    yticklabel pos=left,
    ylabel style={yshift=-3em},xlabel style={yshift=0.3em,align=center},
    yticklabel style={/pgf/number format/fixed,/pgf/number format/fixed},
    legend style={draw=none, line width=1pt, at={(0.5,0.87)}, anchor=south},
    xtick=data,
    axis on top=false,
  ]
\addplot[fill=ugreen!50,draw=ugreen!80, area legend] coordinates { 
    (1,0.939)(2,0.955491947) (3,0.939058123)
};
\addplot[fill=uyellow!50, draw=uyellow!80, area legend] coordinates { 
    (1,0.885313959)(2,0.889536648) (3,0.777767274)
};
\addplot[fill=ured!50, draw=ured!80, area legend] coordinates { 
    (1,1.218757586)(2,1.267692577) (3,1.249981326)
};

\end{axis}

\end{tikzpicture}
}
	\caption{Compare AlpaCaR with baselines, including Alpaca and randomly selected 1k instructions.}
 \label{fig:method_ablation}
\end{figure}

%% file: latex/figures/fig7.tex
\definecolor{color2}{RGB}{098,190,166}
\definecolor{color3}{RGB}{253,186,107}
\definecolor{color4}{RGB}{235,096,070}

\begin{figure}[t!]
\centering
\footnotesize{
\begin{tikzpicture}
    \begin{axis}[
        xbar stacked,
        width=0.5\textwidth,
        height=0.15\textheight,
        nodes near coords,
        nodes near coords align={left},
        point meta=explicit, 
        bar width=1.4em,
        enlarge y limits=0.3,
        xmin=0, 
        xmax=80,
        xtick=\empty,
        xtick pos=bottom,
        ytick={1,2,3},
        ytick = data,
        yticklabels={30B,13B,7B},
        ytick pos=left,
        legend style={at={(0.5,-0.1)}, anchor=north, legend columns=-1, draw=none},
        ]
        \addplot+[xbar,fill=color2!50, draw=color2!80, text=black] plot coordinates {(44,1) [44] (37,2) [37] (46,3) [46]} node[align=center, anchor=center, ] {};
        \addplot+[xbar,fill=color3!50,draw=color3!80, text=black] plot coordinates {(8,1) [8] (15,2) [15] (10,3) [10]};
        \addplot+[xbar,fill=color4!50,draw=color4!80, text=black] plot coordinates {(28,1) [28] (28,2) [28] (24,3) [24]};
        \legend{AlpaCaR win, lose, tie}
    \end{axis}
\end{tikzpicture}
}
\caption{GPT-4 result on Vicuna\_80 dataset: AlpaCaR vs. Alpaca.}
\label{fig:gpt-4_result_vicuna}
\end{figure}

%% file: latex/figures/fig8/fig8.tex
\definecolor{color2}{RGB}{098,190,166}
\definecolor{color3}{RGB}{253,186,107}
\definecolor{color4}{RGB}{235,096,070}
\begin{figure}[t]
\begin{tikzpicture}
\footnotesize{
\begin{axis}[
at={(0,0)},
    ymajorgrids,
    xmajorgrids,
    grid style=dashed,
    width=0.29\textwidth, height=0.22\textwidth,
    legend columns=-1,
    legend entries={Full dataset, select by GPT 3.5, select by CaR},
    legend style={fill opacity=0.5,text opacity =1, draw=none,line width=1pt, at={(0,1.25)}, anchor=north west, nodes={scale=0.8, transform shape} },
    xmin=0.8, xmax=3.2,
    ymin=0.75, ymax=1.05,
    xtick={1,2,3},
    xticklabels = {\scriptsize{LLaMA1}, \scriptsize{LLaMA2}, \scriptsize{LLaMA3} },
    xlabel=\footnotesize{Alpaca 52K},
    xlabel style={yshift=0.1em,align=center},
    scaled ticks=false,
    ]
    \addplot[color2, fill opacity=0.5, mark=*, line width=1pt] file {latex/figures/fig8/dataa/data1.txt}; 
    \addplot[color3, mark=square, line width=1pt] file {latex/figures/fig8/dataa/data2.txt}; 
    \addplot[color4, mark=diamond, line width=1pt] file {latex/figures/fig8/dataa/data3.txt}; 
\end{axis}
}
\hspace{3.8cm}
\footnotesize{
\begin{axis}[        
at={(3.8,0)},
    ymajorgrids,
    xmajorgrids,
    grid style=dashed,
    width=0.29\textwidth, height=0.22\textwidth,
    legend columns=-1,
    xmin=0.8, xmax=3.2,
    ymin=0.65, ymax=0.95,
    xtick={1,2,3},
    xticklabels = {\scriptsize{LLaMA1}, \scriptsize{LLaMA2}, \scriptsize{LLaMA3} },
    xlabel=\footnotesize{Dolly 12K},
    xlabel style={yshift=0.1em,align=center},
    scaled ticks=false,
    ]
    \addplot[color2, fill opacity=0.5, mark=*, line width=1pt] file {latex/figures/fig8/datab/data1.txt}; 
    \addplot[color3, mark=square, line width=1pt] file {latex/figures/fig8/datab/data2.txt}; 
    \addplot[color4, mark=diamond, line width=1pt] file {latex/figures/fig8/datab/data3.txt}; 
\end{axis}
}
\end{tikzpicture}

\caption{Impact of data selection as \textit{pre-trained model become more powerful}.}
\label{fig:pre-trained}
\end{figure}

%% file: latex/figures/fig9/fig9.tex
\definecolor{color2}{RGB}{098,190,166}
\definecolor{color3}{RGB}{253,186,107}
\definecolor{color4}{RGB}{235,096,070}
\begin{figure}[t]
\begin{tikzpicture}
\footnotesize{
\begin{axis}[
at={(0,0)},
    ymajorgrids,
    xmajorgrids,
    grid style=dashed,
    width=0.29\textwidth, height=0.22\textwidth,
    legend columns=-1,
    legend entries={Full dataset, select by GPT 3.5, select by CaR},
    legend style={fill opacity=0.5,text opacity =1, draw=none,line width=1pt, at={(0,1.25)}, anchor=north west, nodes={scale=0.8, transform shape} },
    xmin=0.8, xmax=3.2,
    ymin=0.75, ymax=1.25,
    xtick={1,2,3},
    xticklabels = {\scriptsize{7B}, \scriptsize{13B}, \scriptsize{30B} },
    xlabel=\footnotesize{Model parameter scaling up},
    xlabel style={yshift=0.1em,align=center},
    scaled ticks=false,
    ]
    \addplot[color2, fill opacity=0.5, mark=*, line width=1pt] file {latex/figures/fig9/dataa/data1.txt}; 
    \addplot[color3, mark=square, line width=1pt] file {latex/figures/fig9/dataa/data2.txt}; 
    \addplot[color4, mark=diamond, line width=1pt] file {latex/figures/fig9/dataa/data3.txt}; 
\end{axis}
}
\hspace{3.8cm}
\footnotesize{
\begin{axis}[        
at={(3.8,0)},
    ymajorgrids,
    xmajorgrids,
    grid style=dashed,
    width=0.29\textwidth, height=0.22\textwidth,
    legend columns=-1,
    xmin=0.8, xmax=3.2,
    ymin=1.25, ymax=1.55,
    xtick={1,2,3},
    xticklabels = {\scriptsize{LLaMA1}, \scriptsize{LLaMA2}, \scriptsize{LLaMA3} },
    xlabel=\footnotesize{Trained on alpaca-gpt4},
    xlabel style={yshift=0.1em,align=center},
    scaled ticks=false,
    ]
    \addplot[color2, mark=*, fill opacity=0.5, line width=1pt] file {latex/figures/fig9/datab/data1.txt}; 
    \addplot[color4, mark=square, line width=1pt] file {latex/figures/fig9/datab/data2.txt}; 
\end{axis}
}
\end{tikzpicture}

    \caption{\label{fig:param&data} Impact of data selection as \textit{models parameters} or \textit{instruction quality} increase.}
\end{figure}

%% file: latex/figures/fig11.tex
\definecolor{color2}{RGB}{098,190,166}
\definecolor{color3}{RGB}{253,186,107}
\definecolor{color4}{RGB}{235,096,070}

\begin{figure}[t!]
\centering
\footnotesize{
\begin{tikzpicture}
    \begin{axis}[
        xbar stacked,
        width=0.5\textwidth,
        height=0.15\textheight,
        nodes near coords,
        nodes near coords align={left},
        point meta=explicit, 
        bar width=1.4em,
        enlarge y limits=0.3,
        xmin=0, 
        xmax=150,
        xtick=\empty,
        xtick pos=bottom,
        ytick={1,2,3},
        ytick = data,
        yticklabels={30B,13B,7B},
        ytick pos=left,
        legend style={at={(0.5,0.0)}, anchor=north, legend columns=-1, draw=none},
        ]
        \addplot+[xbar,fill=color2!50, draw=color2!80, text=black] plot coordinates {(55,1) [55] (60,2) [60] (59,3) [59]} node[align=center, anchor=center, ] {};
        \addplot+[xbar,fill=color3!50,draw=color3!80, text=black] plot coordinates {(31,1) [31] (27,2) [27] (40,3) [40]};
        \addplot+[xbar,fill=color4!50,draw=color4!80, text=black] plot coordinates {(64,1) [64] (63,2) [63] (51,3) [51]};
        \legend{AlpaCaR win, lose, tie}
    \end{axis}
\end{tikzpicture}
}
\caption{GPT-4 result on CoachLM\_150 dataset: AlpaCaR vs. Alpaca.}
\label{fig:gpt4_CoachLM}
\end{figure}
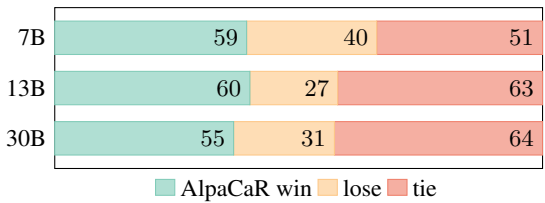

%% file: latex/figures/fig12.tex
\definecolor{color2}{RGB}{098,190,166}
\definecolor{color3}{RGB}{253,186,107}
\definecolor{color4}{RGB}{235,096,070}

\begin{figure}[t!]
\centering
\footnotesize{
\begin{tikzpicture}
    \begin{axis}[
        xbar stacked,
        width=0.5\textwidth,
        height=0.15\textheight,
        nodes near coords,
        nodes near coords align={left},
        point meta=explicit, 
        bar width=1.4em,
        enlarge y limits=0.3,
        xmin=0, 
        xmax=252,
        xtick=\empty,
        xtick pos=bottom,
        ytick={1,2,3},
        ytick = data,
        yticklabels={30B,13B,7B},
        ytick pos=left,
        legend style={at={(0.5,0.0)}, anchor=north, legend columns=-1, draw=none},
        ]
        \addplot+[xbar,fill=color2!50, draw=color2!80, text=black] plot coordinates {(99,1) [99] (102,2) [102] (102,3) [103]} node[align=center, anchor=center, ] {};
        \addplot+[xbar,fill=color3!50,draw=color3!80, text=black] plot coordinates {(56,1) [56] (54,2) [54] (62,3) [62]};
        \addplot+[xbar,fill=color4!50,draw=color4!80, text=black] plot coordinates {(97,1) [97] (96,2) [96] (87,3) [87]};
        \legend{AlpaCaR win, lose, tie}
    \end{axis}
\end{tikzpicture}
}
\caption{GPT-4 result on Self-instruct\_252 dataset: AlpaCaR vs. Alpaca.}
\label{fig:gpt4_Sinstruct}
\end{figure}
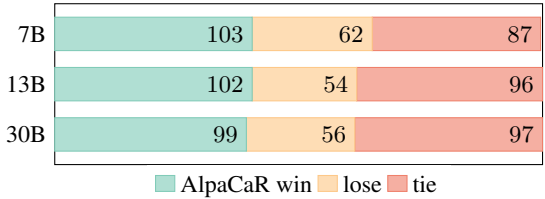

%% file: latex/figures/fig13.tex
\definecolor{color2}{RGB}{098,190,166}
\definecolor{color3}{RGB}{253,186,107}
\definecolor{color4}{RGB}{235,096,070}

\begin{figure}[t!]
\centering
\footnotesize{
\begin{tikzpicture}
    \begin{axis}[
        xbar stacked,
        width=0.5\textwidth,
        height=0.15\textheight,
        nodes near coords,
        nodes near coords align={left},
        point meta=explicit, 
        bar width=1.4em,
        enlarge y limits=0.3,
        xmin=0, 
        xmax=170,
        xtick=\empty,
        xtick pos=bottom,
        ytick={1,2,3},
        ytick = data,
        yticklabels={30B,13B,7B},
        ytick pos=left,
        legend style={at={(0.5,0.0)}, anchor=north, legend columns=-1, draw=none},
        ]
        \addplot+[xbar,fill=color2!50, draw=color2!80, text=black] plot coordinates {(67,1) [67] (73,2) [73] (68,3) [68]} node[align=center, anchor=center, ] {};
        \addplot+[xbar,fill=color3!50,draw=color3!80, text=black] plot coordinates {(44,1) [44] (39,2) [39] (41,3) [41]};
        \addplot+[xbar,fill=color4!50,draw=color4!80, text=black] plot coordinates {(59,1) [59] (58,2) [58] (61,3) [61]};
        \legend{AlpaCaR win, lose, tie}
    \end{axis}
\end{tikzpicture}
}
\caption{GPT-4 result on Pandalm\_170 dataset: AlpaCaR vs. Alpaca.}
\label{fig:gpt4_Pandalm}

\end{figure}
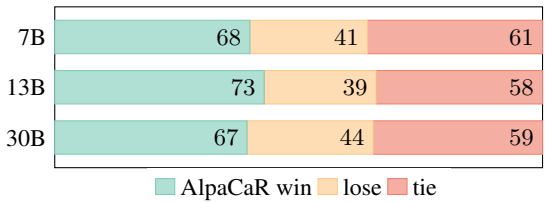